\documentclass[letterpaper]{article} % DO NOT CHANGE THIS
\usepackage{aaai2026}  % DO NOT CHANGE THIS
\usepackage{times}  % DO NOT CHANGE THIS
\usepackage{helvet}  % DO NOT CHANGE THIS
\usepackage{courier}  % DO NOT CHANGE THIS
\usepackage[hyphens]{url}  % DO NOT CHANGE THIS
\usepackage{graphicx} % DO NOT CHANGE THIS
\usepackage{natbib}  % DO NOT CHANGE THIS AND DO NOT ADD ANY OPTIONS TO IT
\usepackage{caption} % DO NOT CHANGE THIS AND DO NOT ADD ANY OPTIONS TO IT
\usepackage{algorithm}
\usepackage{algorithmic}
\usepackage{booktabs}
\usepackage{multirow}
\usepackage{newfloat}
\usepackage{listings}
\usepackage{amsmath, amssymb}
\DeclareCaptionStyle{ruled}{labelfont=normalfont,labelsep=colon,strut=off} % DO NOT CHANGE THIS
\floatstyle{ruled}
\newfloat{listing}{tb}{lst}{}
\floatname{listing}{Listing}
\title{DiagnoLLM: A Hybrid Bayesian Neural Language Framework for Interpretable Disease Diagnosis}
\author{
    Bowen Xu\equalcontrib \textsuperscript{\rm 1},
    Xinyue Zeng\equalcontrib \textsuperscript{\rm 1},\\
    Jiazhen Hu\textsuperscript{\rm 1},
    Tuo Wang\textsuperscript{\rm 1},
    Adithya Kulkarni\textsuperscript{\rm 2}
}
\affiliations{
    \textsuperscript{\rm 1}Department of Computer Science, Virginia Tech\\
    \textsuperscript{\rm 2}Department of Computer Science, Ball State University

}

\usepackage{bibentry}
\begin{document}

\maketitle

\begin{abstract}
Building trustworthy clinical AI systems requires not only accurate predictions but also transparent, biologically grounded explanations. We present \texttt{DiagnoLLM}, a hybrid framework that integrates Bayesian deconvolution, eQTL-guided deep learning, and LLM-based narrative generation for interpretable disease diagnosis. DiagnoLLM begins with GP-unmix, a Gaussian Process-based hierarchical model that infers cell-type-specific gene expression profiles from bulk and single-cell RNA-seq data while modeling biological uncertainty. These features, combined with regulatory priors from eQTL analysis, power a neural classifier that achieves high predictive performance in Alzheimer's Disease (AD) detection (88.0\% accuracy). To support human understanding and trust, we introduce an LLM-based reasoning module that translates model outputs into audience-specific diagnostic reports, grounded in clinical features, attribution signals, and domain knowledge. Human evaluations confirm that these reports are accurate, actionable, and appropriately tailored for both physicians and patients. Our findings show that LLMs, when deployed as post-hoc reasoners rather than end-to-end predictors, can serve as effective communicators within hybrid diagnostic pipelines.
\end{abstract}
\section{Introduction}

Accurate disease diagnosis using transcriptomic data is a central goal in biomedical AI, yet it remains a formidable challenge due to two persistent bottlenecks: (1) the inability to extract cell-type-specific (CTS) signals from noisy bulk RNA-seq data, and (2) the lack of interpretable, clinically actionable explanations for model predictions. These limitations are especially pronounced in neurodegenerative diseases like Alzheimer's Disease (AD), where pathology often manifests in specific brain cell types such as microglia and astrocytes~\cite{blumenfeld2024cell, brendel2022application}. Bulk RNA-seq, the most widely available data modality, aggregates expression over heterogeneous cell populations, thereby masking critical disease signals~\cite{natri2024cell}. Single-cell RNA-seq (scRNA-seq) offers higher resolution~\cite{tasic2018shared, paik2020single, yao2021transcriptomic}, but its high cost, technical complexity, and sparsity in clinical cohorts make it impractical for widespread diagnostic use.

To bridge this gap, deconvolution methods have emerged that estimate CTS profiles from bulk RNA-seq using single-cell references~\cite{xu2025cell,tang2024cell}. However, existing methods often fail to generalize due to their sensitivity to reference-target mismatch, lack of robust uncertainty modeling, and inability to propagate prior biological knowledge~\cite{torroja2019digitaldlsorter}. At the same time, most downstream disease classifiers treat gene expression purely as numerical input, ignoring known regulatory mechanisms such as expression quantitative trait loci (eQTLs) that offer causal, cell-type-aware insights into disease progression~\cite{nica2013expression, natri2024cell}. Even when accurate classifiers are developed, their ``black-box'' nature severely limits clinical adoption. Physicians and patients require not only predictions, but also clear, faithful explanations grounded in known biology~\cite{blumenfeld2024cell}. Recent work has explored the use of large language models (LLMs) in biomedical tasks~\cite{yang2023transformehr,gao2024raw,han2022luna, jiang2025cmfdnet}, yet most applications focus on end-to-end generation or information extraction. These approaches often lack alignment with underlying model behavior, leading to hallucinations and eroding trust in high-stakes settings~\cite{omar2024utilizing,chen2025visrl, zhao2022balf, zhang2024wrong, lin2025plan, zeng2025janusvln, zeng2025future}. Furthermore, LLMs exhibit well-documented limitations in handling symbolic reasoning and numerical precision~\cite{hegselmann2023tabllm, zhang2025vitcot, hu2025flowmaltrans,lu2022understand}, especially when used as primary decision-makers.

\begin{figure}
    \centering
    \includegraphics[width=\linewidth]{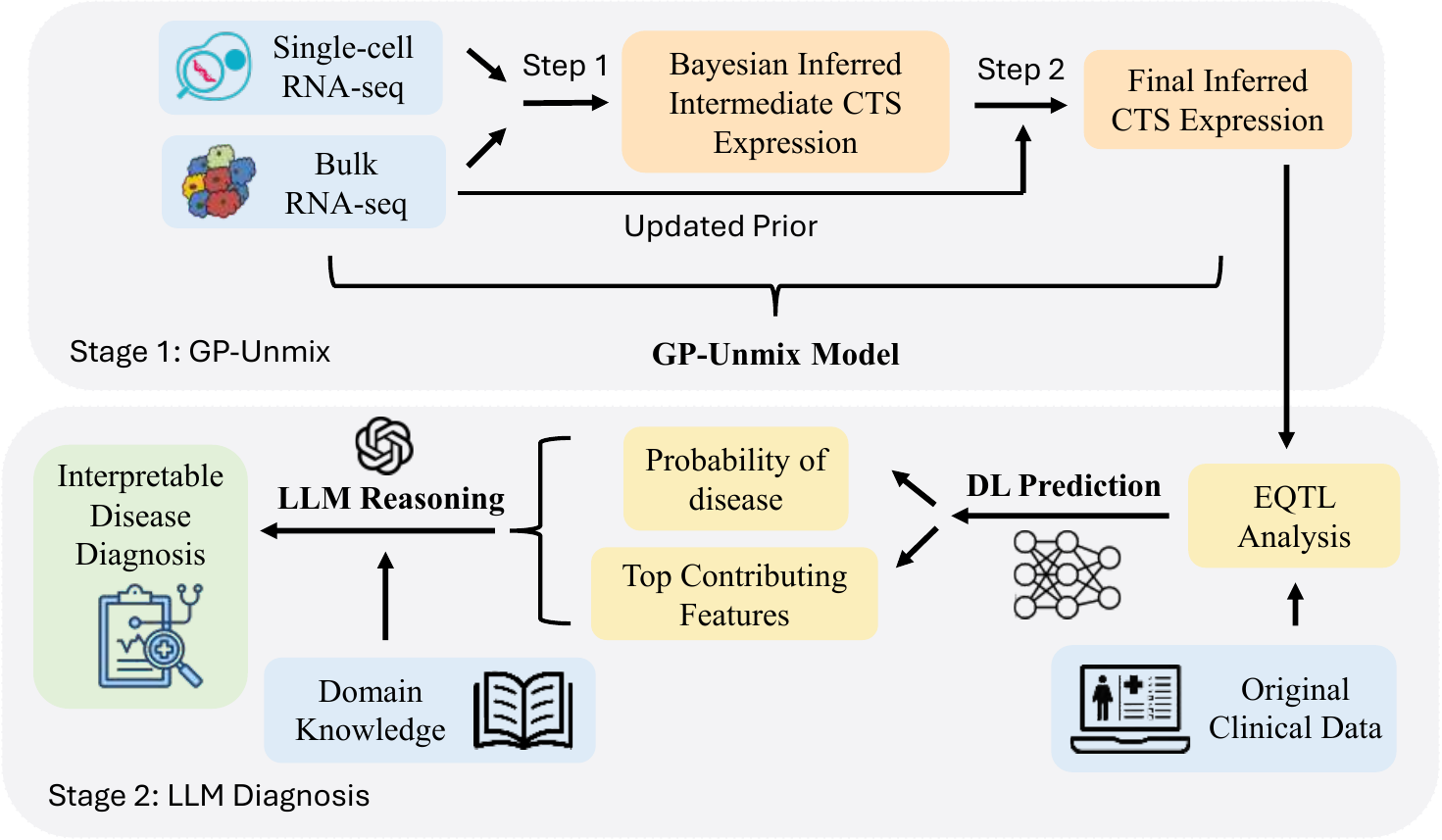}
    \caption{\textbf{Overview of the \textsc{DiagnoLLM} framework.} Stage 1 (\texttt{GP-Unmix}) performs Bayesian deconvolution of bulk RNA-seq into CTS expression using single-cell references. Stage 2 combines eQTL-informed DL predictions with LLM-based reasoning to produce human-readable diagnostic reports, linking model outputs with clinical interpretability.}
    \label{fig:overview}
\end{figure}

We argue that a reliable clinical AI system must unify three capabilities: robust signal extraction, biologically grounded prediction, and audience-specific explanation. To this end, we propose \texttt{DiagnoLLM}, a hybrid neuro-symbolic framework for interpretable disease diagnosis. As shown in Figure~\ref{fig:overview}, DiagnoLLM integrates Bayesian deconvolution, regulatory genomics, and LLM-based explanation into a two-stage diagnostic pipeline. \textbf{Stage 1} introduces \texttt{GP-unmix}, a Gaussian Process-based hierarchical model that deconvolves bulk RNA-seq into CTS expression matrices using single-cell references. GP-unmix introduces posterior refinement steps to correct for reference-target shifts, incorporates uncertainty modeling via multivariate priors, and filters gene-cell-type pairs using biologically informed selection strategies. Compared to existing methods~\cite{tang2024cell, xu2025cell, torroja2019digitaldlsorter}, it achieves significantly higher gene-level recovery across tissues, species, and modalities. \textbf{Stage 2} enriches prediction with \textit{eQTL-derived regulatory features} that guide a two-layer neural network classifier. These regulatory priors emphasize disease-relevant transcriptional mechanisms~\cite{gusev2016integrative,natri2024cell}, improving both classification performance (88.0\% accuracy on AD) and biological alignment. To address the interpretability gap, we introduce a \textit{language-based reasoning module}, where an LLM translates classifier outputs and feature attributions into structured, audience-specific diagnostic reports tailored for clinicians and patients. This module is grounded in attribution scores and domain knowledge, ensuring factual alignment and avoiding free-form hallucination~\cite{yu2025unified}.

We further justify this hybrid design through a targeted divergence analysis. We find that LLMs often misclassify samples with conflicting symbolic cues (e.g., a positive AD label with a negative BETA value), but outperform neural models in out-of-distribution regions by leveraging domain priors. Conversely, the MLP is more robust to statistical variance but fails in rare or extreme cases. This complementary behavior motivates our neuro-symbolic integration: the neural model handles structured prediction, while the LLM serves as a post-hoc communicator and narrative generator.

\noindent Our contributions are threefold:\\
\textbf{1. GP-unmix for cell-type deconvolution:} A Bayesian framework with dynamic prior refinement and gene selection for uncertainty-aware recovery of CTS expression.\\
\textbf{2. eQTL-guided neural prediction:} A classifier trained on deconvolved and regulatory features that demonstrates improved accuracy and mechanistic alignment in AD detection.\\
\textbf{3. LLM-based explanation module:} A structured prompting system that generates clinically actionable reports, evaluated across user trust, rationale completeness, and audience-appropriateness.

DiagnoLLM bridges probabilistic modeling, biological reasoning, and natural language explanation, offering a path forward for interpretable AI in real-world clinical diagnostics.
\section{Related Work}
\label{relatedwork}

\paragraph{CTS Estimation and Deconvolution Models.}
Cell-type-specific (CTS) expression estimation is critical for uncovering disease-relevant signals from heterogeneous tissue data. Traditional deconvolution approaches such as TCA and bMIND~\cite{xu2025cell, torroja2019digitaldlsorter} estimate CTS proportions using single-cell reference data, but often lack uncertainty modeling and fail to generalize across tissues or species. Our framework extends this line of work by introducing \textit{GP-unmix}, a Bayesian model with dynamic posterior refinement and biologically grounded prior selection that produces uncertainty-aware CTS profiles.

\paragraph{LLMs for Disease Analysis.}
Large language models (LLMs) have increasingly been explored for biomedical tasks, including disease classification, literature summarization, and biomarker discovery~\cite{yang2022scbert, levine2024cell2sentence, zeng2025disprotbench}. Understanding disease-relevant cell types is essential for mechanistic modeling in complex conditions like Alzheimer’s Disease~\cite{omar2024utilizing, giuffre2024optimizing}. Advances in single-cell RNA-seq~\cite{miranda2023single, mathys2019single} and CTS analysis~\cite{jagadeesh2022identifying, hampel1998corpus} have enabled high-resolution disease modeling, but extracting insights from such data requires models that are both scalable and interpretable. Most existing LLM-based systems rely on pretraining to encode domain knowledge, but they often operate as black-box classifiers or free-form generators, limiting their reliability in clinical decision-making~\cite{elsborg2023using}.

\paragraph{LLMs for Cell-Type Annotation and Omics Reasoning.}
Recent work has integrated LLMs into single-cell workflows to automate cell-type annotation and improve reproducibility. scInterpreter~\cite{scInterpreter} uses LLMs to interpret scRNA-seq profiles, while the Single-Cell Omics Arena benchmark~\cite{liu2024singlecellomicsarenabenchmark} evaluates LLM performance in multi-omics classification and cross-modality translation. These frameworks demonstrate LLMs' potential to reduce manual burden in biological annotation tasks. However, such systems generally focus on stand-alone annotation rather than aligning explanations with structured model predictions or attribution signals, as we propose in DiagnoLLM.

\paragraph{LLMs for Numerical and Tabular Reasoning.}
LLMs have also been adapted to handle structured numerical and tabular inputs. LIFT~\cite{dinh2022lift} converts structured numerical data into natural language to enable LLMs like GPT-3~\cite{brown2020language} and GPT-J~\cite{gpt-j} to perform classification and regression tasks with performance rivaling or surpassing conventional models. For nonlinear tasks, LIFT outperforms decision trees and deep MLPs, and for regression, it exceeds polynomial and nearest-neighbor methods. TabLLM~\cite{hegselmann2023tabllm} extends this idea by using natural language serialization for zero- and few-shot classification, achieving competitive performance against strong ML baselines including logistic regression, XGBoost~\cite{chen2016xgboost}, LightGBM~\cite{ke2017lightgbm}, TabNet~\cite{arik2021tabnet}, and TabPFN~\cite{hollmann2022tabpfn}, as well as foundation models like T0~\cite{sanh2021multitask}. To improve numeric precision, LUNA~\cite{han2022luna} introduces numeric augmentations and representations (NumTok and NumBed) in transformer models such as BERT~\cite{devlin2018bert} and RoBERTa~\cite{liu2019roberta}. These enhancements allow LLMs to handle numeric inputs more faithfully, but challenges remain regarding symbolic overgeneralization and interpretability in real-world domains~\cite{yang2023transformehr, gao2024raw,kulkarni2025scientific}. In addition to healthcare, LLMs for numerical reasoning have also been explored in finance~\cite{ma2025llm, zhu2024tat} and mathematical modeling~\cite{schwartz2024numerologic, lee2023teaching}, where grounding, consistency, and multi-step reasoning remain open challenges.

Unlike prior methods that employ LLMs as end-to-end predictors or annotation tools, DiagnoLLM adopts a neuro-symbolic architecture that decouples numerical prediction from natural language explanation. The classifier is trained on biologically grounded features derived from Bayesian deconvolution and eQTL-informed priors~\cite{nica2013expression, natri2024cell}, ensuring mechanistic relevance. Meanwhile, the LLM functions as a structured reasoning module, guided by attribution scores and tailored to audience-specific interpretability goals. This hybrid design addresses a central gap in the literature: integrating LLMs into clinical pipelines not as autonomous decision-makers, but as faithful narrators of model behavior. While post-hoc interpretability tools such as SHAP and LIME are commonly used in clinical AI~\cite{lundberg2017unified, ribeiro2016should}, they often lack biological context and are sensitive to perturbation artifacts. In contrast, DiagnoLLM grounds its explanations in both domain knowledge and model saliency, enabling reliable, audience-aware diagnostic narratives.
\section{Methodology}
\label{sec:methodology}

We present \texttt{DiagnoLLM}, a modular framework that combines Bayesian deconvolution, regulatory reasoning, and LLM-based interpretation to enable accurate and interpretable disease diagnosis. It addresses bulk RNA-seq limitations by recovering cell-type-specific signals. We first define the problem, then detail each component.

% In this section, we present \texttt{DiagnoLLM}, a modular framework that integrates Bayesian deconvolution, regulatory reasoning, and large language model-based interpretation for transparent and accurate disease diagnosis. Designed to overcome the limitations of bulk RNA-seq, where gene expression measurements are averaged across heterogeneous cell types, \texttt{DiagnoLLM} disentangles cell-type-specific signals to support both robust prediction and interpretable clinical reporting. We begin by formally defining the problem and then describe each component of the framework in detail.

% This section presents our proposed framework, \texttt{DiagnoLLM}, which integrates statistical deconvolution, genetic regulatory reasoning, and large language model-based interpretation to enable trustworthy and interpretable clinical diagnosis from transcriptomic data. The framework is designed to address the limitations of bulk RNA sequencing (RNA-seq), a technique that measures average gene expression across mixed cell populations, in capturing cell-type-specific expression. By resolving this ambiguity, \texttt{DiagnoLLM} supports both accurate disease classification and human-aligned explanations of underlying biological signals. We begin by formally defining the problem, followed by a modular breakdown of the proposed approach.

\subsection{Problem Statement}

Let $X \in \mathbb{R}^{G \times N}$ denote a bulk RNA-seq gene expression matrix, where $G$ is the number of genes and $N$ is the number of patient samples. Each column $x_i \in \mathbb{R}^G$ represents a bulk gene expression profile that conflates signals from multiple cell types, thereby obscuring disease-relevant biological variation. Our objective is to develop a computational pipeline that, given $X$ and auxiliary metadata (such as genotype and clinical covariates), can: (1) Estimate an uncertainty-aware, cell-type-specific (CTS) expression tensor $Z \in \mathbb{R}^{G \times C \times N}$, where $C$ is the number of cell types. CTS expression refers to the gene activity levels attributable to each individual cell type, which are not directly measurable in bulk RNA-seq, (2) Use these CTS representations in conjunction with expression quantitative trait loci (eQTLs), genetic variants that influence gene expression, to predict disease status (e.g., Alzheimer's Disease), and (3) Produce natural language diagnostic reports that are aligned with model predictions and enriched with clinically relevant biological context, targeted at both physicians and patients.

This problem is especially challenging because (1) ground-truth CTS expression is not observed during training, (2) the influence of genetic variation on expression is often nonlinear and context-dependent, and (3) explanatory outputs must be both accurate and understandable to diverse users.

\subsection{The \textsc{DiagnoLLM} Framework}

To address the limitations of bulk RNA-seq in capturing cell-type-specific signals, the need for biologically grounded prediction, and the demand for interpretable clinical outputs, we propose \textsc{DiagnoLLM}—a modular neuro-symbolic framework that integrates Bayesian deconvolution, regulatory reasoning, and structured language-based interpretation. Unlike prior approaches that treat LLMs as end-to-end predictors, \textsc{DiagnoLLM} separates statistical inference from explanation: a neural classifier predicts disease status using features derived from Bayesian deconvolution and eQTL priors, while an LLM generates post-hoc, audience-specific diagnostic narratives. The framework is guided by three core principles: (1) extracting fine-grained molecular signals from noisy data, (2) grounding predictions in validated regulatory features, and (3) enabling transparent, clinically relevant explanations.

Specifically, DiagnoLLM tackles the three key modeling challenges as follows: (1) To disentangle confounded expression signals, we introduce a Bayesian deconvolution module called \texttt{GP-unmix}, which infers gene-level expression profiles at the resolution of individual cell types. The method leverages sc/snRNA-seq references and employs a two-stage posterior refinement strategy to adapt to domain shifts between reference and target data. (2) To improve predictive accuracy and biological specificity, we incorporate known genetic regulatory variation in the form of eQTL priors. These priors are combined with CTS expression and clinical covariates to train a compact neural classifier that predicts Alzheimer's Disease (AD) status, and (3) To promote trust and usability, we use a large language model (LLM) as a post-hoc reasoning module that generates human-readable diagnostic reports. These reports are grounded in model attributions and tailored for different audiences (e.g., physicians or patients). Figure~\ref{fig:overview} provides an overview of the full pipeline. Below, we describe each of the three components in detail.

\subsubsection{Bayesian Deconvolution via GP-unmix}
\label{sec:gp-unmix}

\paragraph{Motivation.}  
A key obstacle in analyzing bulk RNA-seq is that it conflates gene expression across diverse cell types, thereby masking cell-type-specific (CTS) transcriptional programs that may be crucial for disease mechanisms. For instance, Alzheimer's related dysregulation in astrocytes or microglia may be diluted in bulk measurements dominated by neuronal signatures. Existing deconvolution methods, such as bMIND and TCA~\cite{xu2025cell, torroja2019digitaldlsorter}, focus primarily on estimating cell-type proportions and lack the ability to recover full gene-level CTS expression with robust uncertainty quantification. Moreover, they often fail under domain shifts, such as differences in species, tissue, or sequencing protocols between reference and target datasets.

\texttt{GP-unmix} is a hierarchical Bayesian model designed to address these challenges. It infers CTS expression matrices from bulk data by combining:  
(1) multivariate priors derived from single-cell RNA-seq references,  
(2) posterior refinement to adapt to cohort-specific distributions, and  
(3) a tripartite gene selection strategy to improve stability and biological relevance.  

\paragraph{Model Formulation.}  
Let \(X \in \mathbb{R}^{G \times N}\) denote the bulk RNA-seq matrix of \(G\) genes and \(N\) patient samples. The latent CTS expression tensor is denoted by \(Z \in \mathbb{R}^{G \times C \times N}\), where \(C\) is the number of cell types. For each gene \(g\) and cell type \(j\), we model CTS expression across samples as:

\[
Z_{gj} \sim \mathcal{N}(\mu_j, \Sigma_j), \quad X_i = w_i^\top Z_i + \Gamma_j^\top C_i^{(1)} + w_i^\top B_j C_i^{(2)} + \varepsilon_i
\]

Here, \(w_i \in \mathbb{R}^C\) denotes the cell-type proportions for sample \(i\); \(C_i^{(1)}\) captures bulk-level technical covariates such as batch effects; \(C_i^{(2)}\) represents latent confounders at the cell-type-specific (CTS) level; \(\Gamma_j\) and \(B_j\) are learned adjustment matrices that modulate the influence of these covariates; and \(\varepsilon_i \sim \mathcal{N}(0, \sigma_j^2)\) denotes the observation noise.

\paragraph{Reference-Informed Inference.}  
We initialize priors \((\mu_j, \Sigma_j)\) using mean and covariance statistics from sc/snRNA-seq references (e.g., Tasic~\cite{tasic2018shared}, Yao~\cite{yao2021transcriptomic}). These empirical distributions anchor the generative model and provide biologically grounded starting points.

\paragraph{Posterior Refinement.}  
To accommodate domain shifts between references and target cohorts, we update the priors using posterior samples:

\[
\mu_j^{(2)} \sim \mathcal{N}(\widehat{\mu_j^{(1)}}, \tau^2 I), \quad 
\Sigma_j^{(2)} \sim \text{InvWishart}(\widehat{\Sigma_j^{(1)}}, v)
\]

Inference is performed via MCMC sampling with convergence validated using Gelman–Rubin diagnostics (\(\hat{R} < 1.05\)). We summarize the posterior inference process in Algorithm~\ref{alg:gp-unmix}.

\paragraph{Tripartite Gene Selection.}  
We reduce inference noise by selecting informative gene–cell-type pairs through three filters:
\textit{1. Core markers:} manually curated genes known to mark specific cell types (e.g., \textit{SLC6A12}, \textit{C3}),
\textit{2. Statistical stability:} differentially expressed genes across modalities (\(p_{\text{FDR}} < 0.01\), \(|\log_2 \text{FC}| > 1\)), and
\textit{3. Noise suppression:} filtering based on Seurat-derived differential expression scores. This selection strategy improves downstream CTS estimation, yielding 37–54\% higher Pearson correlation coefficients (PCC) compared to baselines across multiple datasets, especially in low-abundance or high-noise cell populations.

\begin{algorithm}[h!]
\caption{GP-unmix: Posterior Inference}
\label{alg:gp-unmix}
\begin{algorithmic}[1]
\REQUIRE Bulk RNA-seq matrix $X$, reference dataset $\mathcal{R}$, covariates $C^{(1)}$, $C^{(2)}$
\STATE Initialize priors \((\mu_j^{(0)}, \Sigma_j^{(0)})\) from $\mathcal{R}$
\STATE Select gene–cell-type pairs using tripartite gene filtering
\FOR{each pair $(g, j)$}
    \STATE Draw samples: \(Z_{gj}^{(1)} \sim \mathcal{N}(\mu_j^{(0)}, \Sigma_j^{(0)})\)
\ENDFOR
\STATE Fit bulk mixture: \(X_i \approx w_i^\top Z_i + \Gamma_j^\top C_i^{(1)} + w_i^\top B_j C_i^{(2)}\)
\STATE Estimate posteriors \((\widehat{\mu_j^{(1)}}, \widehat{\Sigma_j^{(1)}})\)
\STATE Update priors and repeat MCMC sampling for $T$ iterations
\STATE \textbf{Return} Posterior means \(\mathbb{E}[Z_{gj} \mid X]\), variances \(\text{Var}[Z_{gj} \mid X]\)
\end{algorithmic}
\end{algorithm}

\paragraph{Empirical Validation.}

\begin{figure}[ht]
    \centering
    \includegraphics[width=\linewidth]{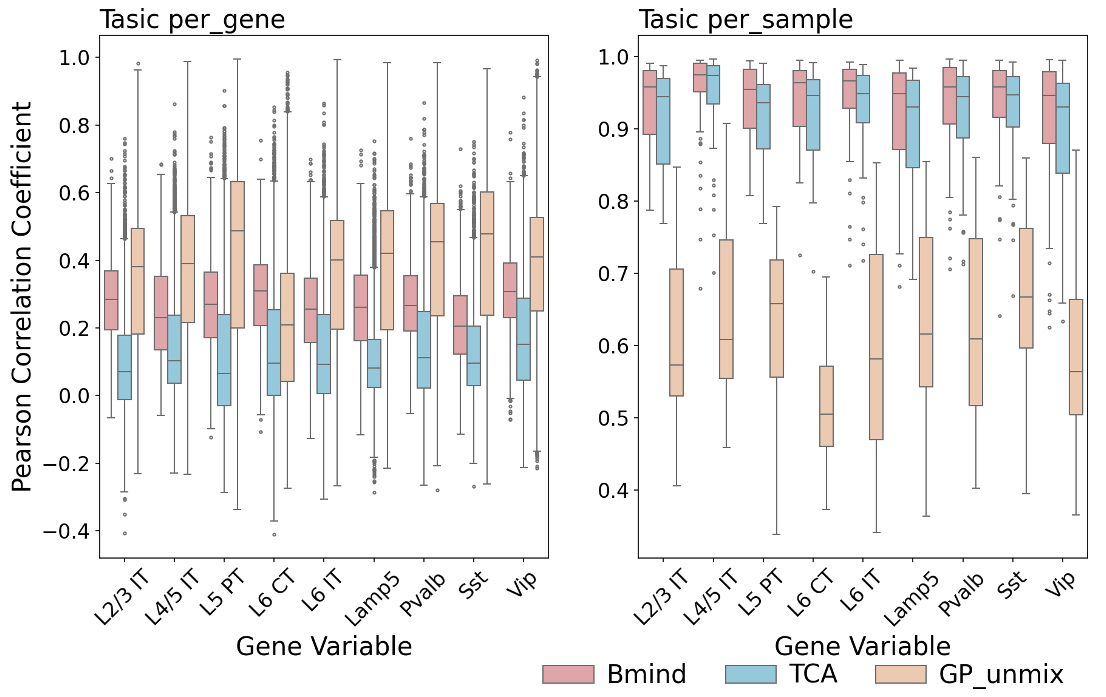}
    \caption{GP-unmix improves CTS recovery over TCA and bMIND across neuron subtypes in the Tasic dataset~\cite{tasic2018shared}.}
    \label{fig:tasic}
\end{figure}

\begin{figure}[ht]
    \centering
    \includegraphics[width=\linewidth]{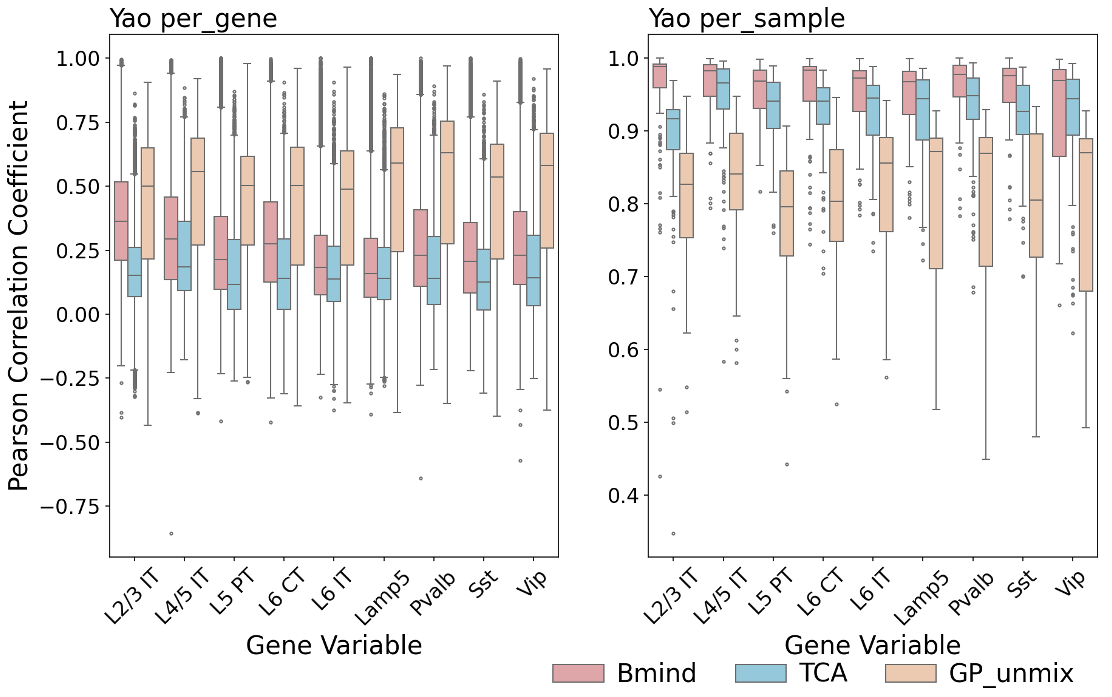}
    \caption{GP-unmix outperforms baselines on astrocytes, microglia, and inhibitory subtypes in the Yao dataset~\cite{yao2021transcriptomic}.}
    \label{fig:yao}
\end{figure}

Figures~\ref{fig:tasic} and~\ref{fig:yao} benchmark GP-unmix on the Tasic and Yao datasets, respectively. It achieves higher Pearson correlation than bMIND and TCA at both per-gene and per-sample levels, particularly on complex and low-abundance populations such as L6 CT, Lamp5, and Pvalb interneurons. In human brain data, GP-unmix achieves a median PCC of 0.82 for microglia and 0.78 for astrocytes. In PBMCs, it yields strong alignment with flow cytometry ground-truth (\(r = 0.71\) for NK cells), outperforming bMIND by over 105\%. These CTS profiles enable downstream regulatory analysis. In the ROSMAP Alzheimer's cohort, GP-unmix reveals astrocyte-linked dysregulation in UDP-glucosyltransferase activity, a pathway implicated in neurodegenerative inflammation, thus demonstrating both predictive utility and biological interpretability.

\subsubsection{eQTL-Guided Disease Classification}

\paragraph{Motivation.}
While CTS expression inferred by GP-unmix provides granular transcriptional features, not all expression variability is biologically meaningful or disease-relevant. Expression quantitative trait loci (eQTLs), genetic variants that modulate gene expression in a cell-type-specific manner, offer mechanistic priors that ground observed expression shifts in underlying regulatory architecture. Incorporating these priors enables the classifier to prioritize biologically plausible features and reduce reliance on noise or confounding signals, particularly in complex disorders like Alzheimer's Disease (AD).

\paragraph{Feature Construction.}
Each patient sample is represented using a concatenation of three biologically motivated components:
\textit{1. CTS expression vectors:} Mean expression values for selected gene–cell-type pairs output by GP-unmix, 
\textit{2. eQTL-derived regulatory priors:} For each gene, we include effect size (\texttt{BETA}), uncertainty (\texttt{SE}), and statistical significance (\texttt{PVAL}), obtained from cohort-specific or public databases, and
\textit{3. Covariates:} Demographic and technical confounders such as age, sex, and batch indicators. These vectors are standardized and projected into a compact, biologically grounded feature space for classification.

\paragraph{Classifier Design.}
We implement a two-layer feedforward neural network (MLP) trained using binary cross-entropy loss. The model includes ReLU activations, dropout regularization, and early stopping based on validation loss. The simplicity of this architecture is intentional to ensure transparency and compatibility with gradient-based attribution methods.

\paragraph{Attribution and Interpretability.}
To enable post-hoc explanation via LLMs, we compute feature attributions using Integrated Gradients. These attributions, indicating the contribution of each input feature to the model's output, serve as intermediate representations for structured diagnostic narrative generation in the next stage of the pipeline.

\subsubsection{LLM-Based Diagnostic Interpretation}
\label{sec:llm}

\paragraph{Motivation.}
Despite advances in disease classification accuracy, clinical deployment requires models to produce explanations that are understandable, faithful, and tailored to different stakeholders. LLMs offer a promising solution by translating structured model outputs into human-readable reports. However, when used as standalone predictors, LLMs often lack numerical grounding and exhibit inconsistency in reasoning. We address this by using the LLM as a post-hoc interpretability module, conditioned on model outputs, feature attributions, and biomedical priors, to generate explanations aligned with both statistical evidence and clinical knowledge.

\paragraph{Input Representation.}
The LLM receives as input a structured prompt composed of:
\textit{1. Predicted label and confidence:} Binary AD prediction and associated softmax probability from the classifier,
\textit{2. Feature attributions:} Saliency scores from Integrated Gradients, reflecting which CTS expression and eQTL features influenced the model’s decision, and
\textit{3. Biological priors:} Contextual information on key genes (e.g., \textit{APOE}, \textit{TREM2}) derived from public literature or cohort-specific findings. These components are serialized into natural language using prompt templates, enabling compatibility with standard instruction-tuned LLMs such as GPT-3.5 or BioMedLM.

\paragraph{Prompting Strategies.}
We experiment with three progressively structured prompting approaches:
\textit{1. Direct Reasoning:} The LLM receives a flattened list of features and is asked to output a diagnosis label with minimal guidance.
\textit{2. Step-by-Step Reasoning:} The LLM first summarizes distributions of features across known AD and non-AD populations, then compares the test case against these learned distributions before making a decision.
\textit{3. Step-by-Step with Domain Knowledge:} The prompt is augmented with biomedical background (e.g., known AD gene signatures), encouraging the LLM to reason using established clinical knowledge alongside the model's attributions.

\paragraph{Audience-Specific Report Generation.}
Using the same internal representation, the LLM is prompted to generate diagnostic narratives tailored to two user groups:
\textit{1. Clinician reports:} Emphasize biomarker relevance, statistical confidence, and biological pathways involved, and
\textit{2. Patient summaries:} Simplify terminology and focus on actionable insights while preserving factual fidelity. These reports help close the loop between high-performance predictive modeling and real-world interpretability demands in precision medicine.

\subsection{LLM as Post-hoc Interpretability Module}

While LLMs can occasionally outperform deep learning models in low-data settings, their use as end-to-end classifiers in clinical contexts is risky due to sensitivity to heuristics and lack of numerical rigor. In \texttt{DiagnoLLM}, we instead deploy the LLM as a post-hoc interpretability module, conditioned on outputs from a statistically trained neural model. This design ensures predictive stability while enabling transparent and audience-aligned explanation.

Each LLM-generated report is prompted with: (a) the MLP-predicted probability $p(\mathrm{AD})$, and (b) the top-5 most influential features determined by Integrated Gradients, including feature values, attribution scores, and reference ranges. The LLM is then prompted to output: (1) a binary decision (yes/no for AD), (2) a rationale grounded in feature-level reasoning, and (3) next-step recommendations (e.g., clinical follow-up, lifestyle guidance). We generate two versions per report: one for clinicians (technical terms and differential risk framing) and one for patients (plain language with actionable summaries).

\section{Experiments}
\label{sec:experiments}

We evaluate \textsc{DiagnoLLM}'s hybrid architecture across three fronts: (1) LLM prompting vs. deep learning baselines, (2) divergence analysis of LLM and MLP behaviors, and (3) expert assessments of explanation quality.

% This section presents a comprehensive evaluation of \textsc{DiagnoLLM}, demonstrating how its hybrid architecture effectively combines predictive accuracy with interpretability. We first benchmark LLM prompting strategies against a deep learning baseline under different data regimes, then conduct a targeted divergence analysis to motivate the complementary roles of LLMs and MLPs. Finally, we evaluate our LLM-based explanation module through expert-rated user studies and case analyses. 

\subsection{Classification Setup and Baseline Comparison}

We evaluate LLM prompting strategies for binary Alzheimer's Disease (AD) classification on a structured clinical dataset containing 28 features per patient, including biomarkers, vitals, and genetic markers. Two training regimes are considered: \textit{low-data} (50 training samples) and \textit{full-data} (100 training samples), both evaluated on a fixed held-out test set of 100 randomly-sampled samples to ensure consistent comparison.

As a structured baseline, we implement a two-layer multilayer perceptron (MLP) trained on the same feature set. The MLP comprises two fully connected layers (input $\rightarrow$ 16 $\rightarrow$ 8 $\rightarrow$ 1) with ReLU activations, followed by a sigmoid output for binary prediction. The model is trained using the Adam optimizer (learning rate = 0.001) with binary cross-entropy (BCE) loss. The LLM used in our study is GPT-4o-mini. 
% For each training regime, class-balanced subsets are randomly sampled to maintain label parity.

Table~\ref{tab:prompt-results} summarizes classification performance across all methods. The MLP achieves the highest accuracy in the full-data setting, but underperforms the LLM in the low-data regime when domain knowledge is incorporated. The structured prompt with domain guidance (\textbf{LLM+Domain}) achieves 90\% accuracy and 0.89 F1, surpassing the MLP by 3 points, underscoring the utility of knowledge-informed prompting in data-scarce medical contexts.

\begin{table}[h]
\centering
\begin{tabular}{lcc|cc}
\toprule
\multirow{2}{*}{\textbf{Method}} & \multicolumn{2}{c|}{\textbf{Train Size = 100}} & \multicolumn{2}{c}{\textbf{Train Size = 50}} \\
 & \textbf{ACC} & \textbf{F1} & \textbf{ACC} & \textbf{F1} \\
\midrule
MLP (DL)       & \textbf{0.88} & \textbf{0.86} & 0.87 & 0.88 \\
LLM-Direct     & 0.48 & 0.43 & 0.50 & 0.49 \\
LLM-Step       & 0.70 & 0.62 & 0.77 & 0.75 \\
LLM+Domain     & 0.74 & 0.69 & \textbf{0.90} & \textbf{0.89} \\
\bottomrule
\end{tabular}
\caption{Classification accuracy and F1 across prompting variants and MLP baseline under different data regimes.}
\label{tab:prompt-results}
\end{table}

\subsection{Divergence Analysis of LLM and MLP Reasoning}
\label{sec:divergence_analysis}

While our quantitative results highlight the benefits of combining deep learning with structured prompting, this section offers deeper insight into \texttt{DiagnoLLM}'s hybrid design. Specifically, we compare the behavioral failure modes of the MLP and LLM components to justify why we treat the LLM as a post-hoc reasoner rather than a standalone classifier.

\subsubsection{Symbol-Sensitive Failures in LLMs}

LLMs tend to apply overly rigid rules when interpreting symbolic patterns, such as associating the sign of BETA with class labels. We curated a test subset of 100 Alzheimer's-positive (AD) samples with negative BETA values, along with corresponding SE and PVAL\footnote{BETA denotes the effect size of an eQTL on gene expression, SE is its standard error, and PVAL indicates the associated statistical significance.}, to evaluate LLM behavior under symbolic conflicts. On this diagnostic subset, the LLM achieved only \textit{38.71\%} accuracy, compared to the MLP's \textit{89.19\%}. Manual inspection revealed that the LLM often ignored the uncertainty (SE) or statistical significance (PVAL), leading to heuristic overfitting.

\subsubsection{Out-of-Distribution Magnitude Failures in MLPs}

In contrast, the MLP underperforms on test samples with feature values outside the training distribution. We constructed a subset of 100 such outlier instances with at least one feature exceeding $\pm1\sigma$ from the training mean. On this set, the MLP accuracy dropped to \textit{61.26\%}, while the LLM, leveraging biomedical priors, reached \textit{88.00\%}. These results underscore the MLP's fragility in sparse data regimes and the LLM's generalization via conceptual knowledge.

\subsubsection{Case Studies of Model Divergence}

To illustrate the above patterns, we present three representative cases in Table~\ref{tab:divergence-cases}. Each example reveals a distinct strength or failure mode and supports the use of a hybrid structure combining MLP precision with LLM transparency.

\begin{table}[h]
\centering
\small
\begin{tabular}{p{0.6cm} p{1.8cm} p{0.4cm} p{0.4cm} p{0.4cm} p{2.3cm}}
\hline
\textbf{Case} & \textbf{Features} & \textbf{Label} & \textbf{MLP} & \textbf{LLM} & \textbf{Key Insight} \\
\hline
1: LLM Error & BETA = -0.03185; SE = 0.04911; PVAL = 0.51671 & AD & AD & non-AD & LLM applies rigid rule to BETA sign; ignores uncertainty and p-value. \\
\hline
2: MLP Error & BETA = 0.976; SE = 0.571; PVAL = 2.166 & AD & non-AD & AD & MLP misclassifies due to extreme input; LLM leverages domain priors. \\
\hline
3: Agreement & BETA = 0.041; SE = 0.061; PVAL = 0.436 & AD & AD & AD & LLM offers explicit rationale tied to BETA range; MLP is opaque. \\
\hline
\end{tabular}
\caption{Examples of divergence and complementarity between MLP and LLM in AD classification.}
\label{tab:divergence-cases}
\end{table}

These results validate our architectural choice: the MLP provides stable decision-making grounded in statistical learning, while the LLM augments transparency and robustness via biomedical priors. Used as a post-hoc reasoning engine, the LLM mitigates model brittleness and enables explanatory alignment with human reasoning, critical for real-world adoption in clinical settings.

\subsection{LLM-Guided Interpretation: Case Studies and Evaluation}

While LLMs have shown promise in low-data regimes, their use as standalone clinical classifiers remains limited by symbolic rigidity and poor numerical grounding. \texttt{DiagnoLLM} instead adopts a hybrid strategy: the LLM functions purely as a post-hoc interpretability module, translating neural model outputs into structured, audience-specific diagnostic narratives. This separation ensures predictive reliability while enhancing transparency through saliency-aware, domain-grounded explanation.

We now present two representative case studies to illustrate how the LLM explanations reflect biological plausibility and align with known Alzheimer’s Disease (AD) mechanisms.

\paragraph{Patient A: High-Risk APOE Carrier}

% \vspace{-0.5em}
\begin{itemize}
    \item \textbf{Summary:} APOE-positive, $p(\mathrm{AD}) = 0.83$, Final decision: AD
    \item \textbf{Key Features:} Elevated triglycerides (220.78 mg/dL), low albumin, poor diet, and high creatinine—all well-established risk factors amplified in APOE carriers~\cite{hunsberger2019role}.
    \item \textbf{Model Alignment:} MLP prioritizes lipids and inflammation; LLM weaves these into a narrative with recommendations for lipid management and neuroimaging.
\end{itemize}

\paragraph{Patient B: Low-Risk APOE Non-Carrier}

% \vspace{-0.5em}
\begin{itemize}
    \item \textbf{Summary:} APOE-negative, $p(\mathrm{AD}) = 0.10$, Final decision: non-AD
    \item \textbf{Key Features:} Age (80.37), elevated homocysteine (16.26 $\mu$mol/L), LDL cholesterol (111.93 mg/dL), and short sleep duration (5 hours).
    \item \textbf{Model Alignment:} MLP flags vascular and metabolic markers with moderate weights. LLM offers a conservative explanation, advising monitoring and lifestyle adjustments.
\end{itemize}

These examples demonstrate the explanatory alignment between the LLM outputs and model attributions, as well as their grounding in established AD biology. They highlight the utility of \texttt{DiagnoLLM} in producing audience-specific, biologically coherent diagnostic reports that support real-world clinical communication.

\subsection{Simulated User Study on Trust and Explanation Quality}
\label{sec:user_study}

To assess the real-world utility of our interpretability module, we conducted a simulated user study focused on how structured, audience-specific LLM-generated reports are perceived by human experts. The study involved raters with clinical training, who evaluated the trustworthiness, clarity, and actionability of diagnostic reports across two audience types: physicians and laypersons.

\subsubsection{Evaluation Design}
We sampled 60 explanation reports (30 clinician-facing and 30 patient-facing) and asked three expert raters to independently score each report across five qualitative dimensions:
\textit{1. Prediction Agreement:} Does the rater agree with the model’s decision?
\textit{2. Feature Rationale:} Are key predictive features explicitly referenced?
\textit{3. Actionability:} Are meaningful next steps (e.g., lifestyle changes, follow-ups) suggested?
\textit{4. Justification Coherence:} Is the explanation logically connected to the outcome?
\textit{5. Stylistic Fit:} Is the tone appropriate for the target audience?

\subsubsection{Results Summary}

\begin{table}[h]
\centering
\begin{tabular}{p{2cm} p{1.3cm} p{3.7cm}}
\toprule
\textbf{Evaluation Dimension} & \textbf{Agreement (\%)} & \textbf{Comment} \\
\midrule
Prediction Agreement & 100.0 & Raters accepted all decisions \\
Feature Rationale & 45.0 & Attribution often underspecified \\
Actionability & 94.2 & Strong next-step guidance \\
Justification Coherence & 17.5 & Logical flow needs refinement \\
Stylistic Fit & 76.7 & Generally well-matched tone \\
\bottomrule
\end{tabular}
\caption{Expert ratings across explanation dimensions (n=60 reports).}
\label{tab:user_study}
\end{table}

% The results in Table~\ref{tab:user_study} show that while agreement with model decisions was universal, only 45\% of reports referenced the most important features, and fewer than 20\% provided a clear reasoning path from evidence to conclusion. Nonetheless, reports were overwhelmingly deemed actionable (94.2\%) and appropriately styled for their intended audience (76.7\%).
As Table~\ref{tab:user_study} shows, all reports aligned with model decisions, but only 45\% referenced key features, and fewer than 20\% offered clear evidence-to-conclusion reasoning. Still, 94.2\% were rated actionable, and 76.7\% were appropriately styled for their target audience.

\subsubsection{Illustrative Examples}

\begin{itemize}
\item \textbf{Patient A (Physician Report):} Correct decision, but cited only ``age'' as rationale despite stronger drivers. Justification was vague, but next steps and tone were appropriate.
\item \textbf{Patient B (Layperson Report):} Correct diagnosis, but explanation contained technical terms (``LDL,'' ``homocysteine''), making it less accessible.
\item \textbf{Patient C (Physician Report):} Strong alignment across all dimensions—clear reasoning, feature-based rationale, appropriate language.
\end{itemize}

These findings confirm that \textsc{DiagnoLLM} produces actionable, trusted explanations, but also reveal weaknesses in feature attribution and logical coherence. These results underscore the need for stronger prompt grounding and closer integration between model interpretation and generation.
% These findings validate \textsc{DiagnoLLM}'s ability to generate trusted and actionable diagnostic narratives, but also highlight areas for improvement. The low rate of feature attribution and limited logical justifications suggest that prompt design could benefit from stronger grounding in model saliency signals. This evaluation reinforces the importance of aligning explanatory content with user expectations in high-stakes domains and motivates future work on tighter coupling between model interpretation and generation modules.

\section{Conclusion}

We present \textsc{DiagnoLLM}, a modular diagnostic framework that unifies Bayesian deconvolution, genetic regulatory modeling, and LLM-based interpretability to address key challenges in clinically grounded, cell-type-aware disease prediction. Our proposed \textit{GP-unmix} model recovers uncertainty-aware, cell-type-specific gene expression with high fidelity, substantially outperforming existing deconvolution approaches across species and tissues. By integrating these expression profiles with eQTL priors, a deep learning classifier, and structured language model prompts, \textsc{DiagnoLLM} achieves robust prediction performance alongside transparent, audience-specific explanations. Through detailed divergence analysis, biological case studies, and a simulated user study, we demonstrate that our hybrid design mitigates the symbolic rigidity of LLMs and the numerical fragility of neural predictors, offering interpretability without sacrificing performance.

\bibliography{aaai2026}

% % Check whether the conference requires a reproducibility checklist to be included in the paper.
% % If so, you can uncomment the following line and ajust the path to include it.
%\input{./ReproducibilityChecklist/LaTeX/ReproducibilityChecklist.tex}

\end{document}